\documentclass[conference]{IEEEtran}

\usepackage[utf8]{inputenc}
\usepackage[T1]{fontenc}
\usepackage{microtype}
\usepackage[ngerman,english]{babel}
\usepackage{listings} 
\usepackage{amsmath}
\usepackage[hidelinks]{hyperref}
\usepackage{cleveref}
\usepackage{multicol}
\usepackage{glossaries}
\usepackage[commandnameprefix=always,todonotes={textsize=tiny,obeyFinal}]{changes}
\usepackage{tikz}
\usepackage[binary-units=true,per-mode=repeated-symbol]{siunitx}
\usepackage{dblfloatfix}
\usepackage{booktabs}
\usepackage[frozencache=true,cachedir=.]{minted} 

\definecolor{color0}{HTML}{1f77b4}  
\definecolor{color1}{HTML}{ff7f0e}  
\definecolor{color2}{HTML}{2ca02c}  
\definecolor{color3}{HTML}{d62728}  
\definecolor{color4}{HTML}{9467bd}  
\definecolor{color5}{HTML}{8c564b}  
\definecolor{color6}{HTML}{e377c2}  
\definecolor{color7}{HTML}{7f7f7f}  
\definecolor{color8}{HTML}{bcbd22}  
\definecolor{color9}{HTML}{17becf}  

\usetikzlibrary{calc,backgrounds,fit}
\usetikzlibrary{arrows,arrows.meta,bending,positioning,calc,patterns,backgrounds,fit,decorations.pathreplacing,shapes.geometric}
\setminted{fontsize=\footnotesize,bgcolor=gray!10}
\setmintedinline{bgcolor=white!0}

\usepackage[backend=biber,maxbibnames=1,minbibnames=1,mincitenames=1,firstinits=true,sorting=none,style=ieee,doi=true,eprint=false,isbn=false,url=false]{biblatex}
\newacronym{adc}{ADC}{analog-to-digital converter}
\newacronym{adex}{AdEx}{adaptive exponential integrate-and-fire}
\newacronym{afib}{AF}{atrial fibrillation}
\newacronym{ann}{ANN}{artificial neural network}
\newacronym{asic}{ASIC}{application-specific integrated circuit}
\newacronym{asicab}{\acrshort{asic} adapter \acrshort{pcb}}{\acrlong{asic} adapter \acrlong{pcb}}
\newacronym{api}{API}{application programming interface}
\newacronym{bmbf}{BMBF}{German Federal Ministry of Education and Research}
\newacronym{bptt}{BPTT}{backpropagation through time}
\newacronym{bram}{BRAM}{block random-access memory}
\newacronym{bss2}{\mbox{BSS-2}}{Brain\mbox{ScaleS-2}}
\newacronym{bss1}{\mbox{BSS-1}}{Brain\mbox{ScaleS-1}}
\newacronym{bss2os}{\gls{bss2} OS}{\gls{bss2} Operating System}
\newacronym{bss}{BSS}{BrainScaleS}
\newacronym{cdnn}{CDNN}{convolutional deep neural network}
\newacronym{cpu}{CPU}{central processing unit}
\newacronym{dfki}{DFKI}{German Research Centre for Artificial Intelligence}
\newacronym{dma}{DMA}{direct memory access}
\newacronym{dram}{DRAM}{dynamic random-access memory}
\newacronym{ecg}{ECG}{electrocardiogram}
\newacronym{fpga}{FPGA}{field-programmable gate array}
\newacronym{gbe}{GbE}{gigabit ethernet}
\newacronym{i2c}{I\textsuperscript{2}C}{Inter-Integrated Circuit}
\newacronym{ic}{IC}{integrated circuit}
\newacronym{isa}{ISA}{instruction set architecture}
\newacronym{itl}{ITL}{in-the-loop}
\newacronym{jit}{JIT}{just-in-time}
\newacronym{lvds}{LVDS}{low-voltage differential signaling}
\newacronym{lif}{LIF}{leaky-integrate and fire}
\newacronym{li}{LI}{leaky integrator}
\newacronym{mac}{MAC}{multiply–accumulate}
\newacronym{madc}{MADC}{membrane \acrshort{adc}}
\newacronym{mse}{MSE}{mean squared error}
\newacronym{cadc}{CADC}{columnar \acrshort{adc}}
\newacronym{pcb}{PCB}{printed circuit board}
\newacronym{ppu}{\acrshort{simd} \acrshort{cpu}}{\acrlong{simd} \acrlong{cpu}}
\newacronym{relu}{ReLU}{rectified linear unit}
\newacronym{rtl}{RTL}{Register Transfer Level}
\newacronym{gd}{GD}{gradient descent}
\newacronym{simd}{SIMD}{single instruction, multiple data}
\newacronym{snn}{SNN}{spiking neural network}
\newacronym{sodimm}{\mbox{SO-DIMM}}{small outline dual in-line memory module}
\newacronym{sram}{SRAM}{static random-access memory}
\newacronym{stdp}{STDP}{spike timing dependent plasticity}
\newacronym{stp}{STP}{short term plasticity}
\newacronym{rnn}{RNN}{recurrent neural network}
\newacronym{rsnn}{RSNN}{recurrent spiking neural network}
\newacronym{nasprop}{NASProp}{neuromorphic accumulative spike propagation}
\newacronym{vu}{VU}{vector unit}
\newacronym{udp}{UDP}{user datagram protocol}
\newacronym{cd}{CD}{continuous deployment}
\newacronym{ci}{CI}{continuous integration}
\newacronym{hpc}{HPC}{high-performance computing}
\newacronym{gpu}{GPU}{graphics processing unit}
\newacronym{usb}{USB}{universal serial bus}

\begin{document}

\title{Integrating programmable plasticity in experiment descriptions for analog neuromorphic hardware}

\author{%
\IEEEauthorblockN{Philipp Spilger}%
\IEEEauthorblockA{%
	Kirchhoff Institute for Physics\\
	Heidelberg University, Germany\\
	Email: pspilger@kip.uni-heidelberg.de
}%
\and%
\IEEEauthorblockN{Eric Müller}%
\IEEEauthorblockA{%
	Kirchhoff Institute for Physics\\
	Heidelberg University, Germany\\
	Email: mueller@kip.uni-heidelberg.de
}
\and%
\IEEEauthorblockN{Johannes Schemmel}%
\IEEEauthorblockA{%
	Kirchhoff Institute for Physics\\
	Heidelberg University, Germany
}%
}

\maketitle

\begin{abstract}
The study of plasticity in \acrlongpl{snn} is an active area of research.
However, simulations that involve complex plasticity rules, dense connectivity/high synapse counts, complex neuron morphologies, or extended simulation times can be computationally demanding.
The \acrlong{bss2} neuromorphic architecture has been designed to address this challenge by supporting ``hybrid'' plasticity, which combines the concepts of programmability and inherently parallel emulation.
In particular, observables that are expensive in numerical simulation, such as per-synapse correlation measurements, are implemented directly in the synapse circuits.
The evaluation of the observables, the decision to perform an update, and the magnitude of an update, are all conducted in a conventional program that runs simultaneously with the analog neural network.
Consequently, these systems can offer a scalable and flexible solution in such cases.
While previous work on the platform has already reported on the use of different kinds of plasticity, the descriptions for the \acrlong{snn} experiment topology and protocol, and the plasticity algorithm have not been connected.
In this work, we introduce an integrated framework for describing \acrlong{snn} experiments and plasticity rules in a unified high-level experiment description language for the \acrlong{bss2} platform and demonstrate its use.

\end{abstract}

\begin{IEEEkeywords}
hardware abstraction, plasticity, online learning, neuromorphic
\end{IEEEkeywords}

\section{Introduction}\label{sec:introduction}

In spiking neural network simulations, plasticity mechanisms can significantly increase computational load, especially since the evaluation of plasticity effects often requires longer simulation times~\cite{jordan2018extremely}.
Neuromorphic hardware offers a potential solution by providing more scalable alternatives:
Programmable digital neuromorphic systems, such as SpiNNaker~\cite{furber2013overview,rhodes2018spynnaker} or Loihi~\cite{orchard2021efficient}, while flexible, still rely on the same numerical algorithms;
however, they may offer more efficient simulations through asynchronous, event-driven operation---a feature difficult to achieve on conventional \gls{hpc} machines used for \gls{snn} simulations.
Some analog neuromorphic architectures integrate analog sensors (e.g., for measuring pre-/post-synaptic correlation in the synapse circuit) with digital implementations of the weight update algorithm.
While early systems lacked online reconfigurability for synaptic connectivity/structural plasticity or the algorithm itself, more recent developments allow for online reconfiguration of the connectome and even adjustments to the learning rules during the network operation~\cite{thakur2018mimicthebrain_nourl}.
Simulators on the other hand---due to their ability to interrupt the temporal evolution of the network dynamics---easily support plasticity by adding numerical computations in network nodes, e.g., neurons in \texttt{nest}~\cite{gewaltig2007nest}, or additional equations, e.g., in \texttt{brian2}~\cite{stimberg2019brian}.

Applications can include plasticity models that closely mimic biological archetypes~\cite{mateos2019impact,yger2015models,luboeinski2021memory}, as well as research into meta-level concepts such as evolutionary plasticity~\cite{jordan2021evolving} or learning-to-learn~\cite{bohnstingl2019neuromorphic} approaches.

In recent years, code generation has become a common tool for software simulators to enable the automatic translation from high-level descriptions into efficient code~\cite{blundell2018code}:
while some simulators directly target \gls{gpu} \glspl{api}~\cite{yavuz2016genn}, others focus on intermediate numerical representations, e.g., \cite{sabne2020xla}, that can be compiled to different numerical accelerators~\cite{manna2023frameworks}.

Recent work on the \gls{bss2} hardware already demonstrated various kinds of plasticity~\cite{billaudelle2021structural,billaudelle2020versatile,bohnstingl2019neuromorphic,cramer2020control,cramer2023autocorrelations,wunderlich2019demonstrating}.
However, there was no overarching concept that could be used to express both the network description and the plasticity rule in a unified manner.
In this work, we build on the \gls{bss2} software stack~\cite{mueller2022scalable} to provide a PyNN-based~\cite{davison08pynn} modeling concept of network topology, programmable plasticity and experiment protocol.

\section{Methods}\label{sec:methods}

\begin{figure}
	\includegraphics[width=\columnwidth]{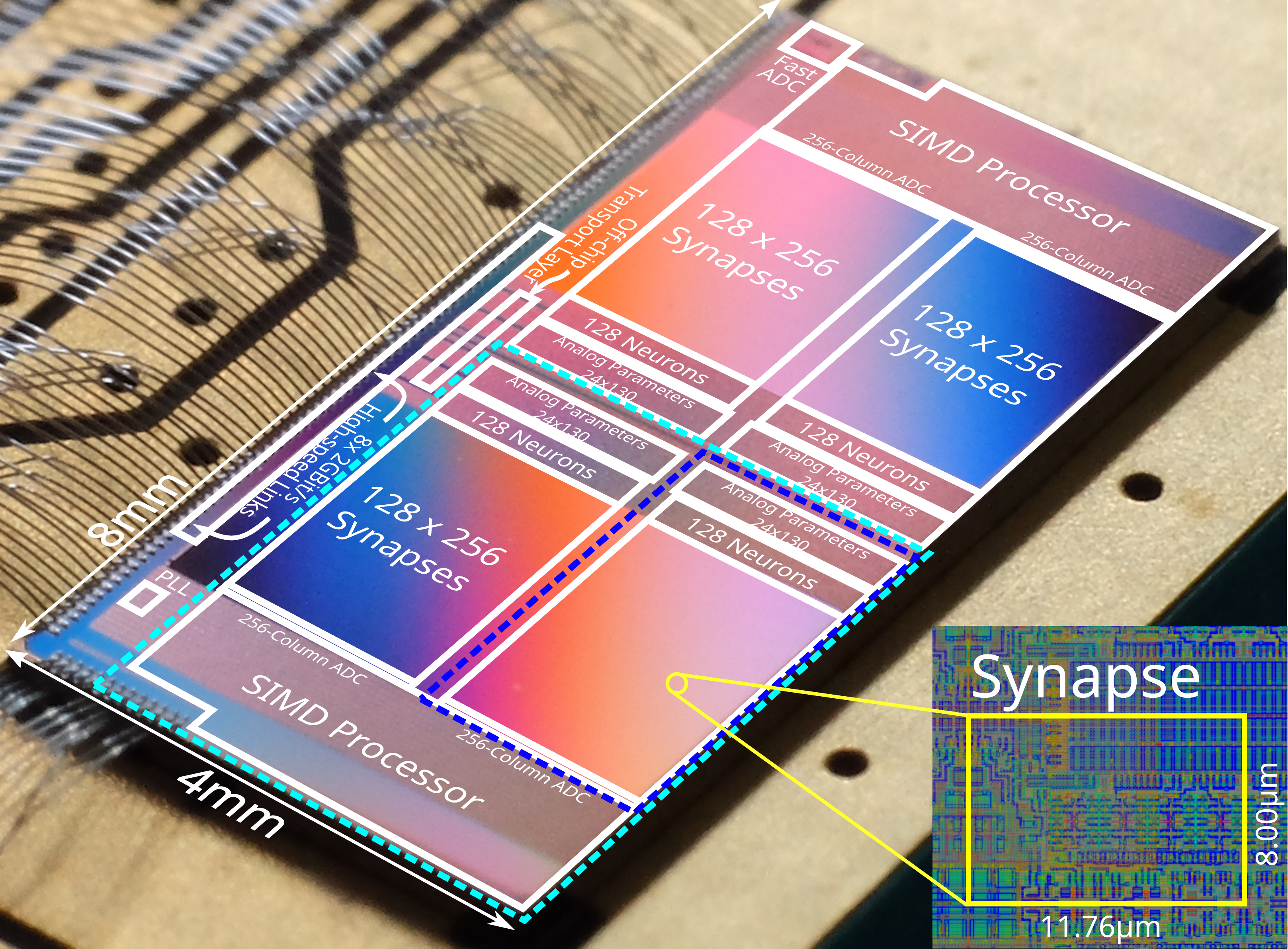}
	\caption{%
		\label{fig:methods:chipfoto}
		A photo of the \gls{bss2} chip with its schematic on top.
		One embedded \gls{simd} processor per chip half accesses on-chip observables, such as synapse-local pre-/post-timed spike correlation information, mean neuronal activities, or firing rates, and modifies neuron and synapse parameterization (e.g., weights), and topology (e.g., to implement structural plasticity) defined by a plasticity algorithm.
	}
\end{figure}

\Gls{bss2} is a mixed-signal neuromorphic hardware platform with tunable, typically 1000-fold accelerated network dynamics compared to biological counterparts~\cite{pehle2022brainscales2}.
Its neural network core features \num{512} analog neuron circuits implementing the \gls{adex} neuron model with \num{256} synapses each per chip.
Digital events are generated by on-chip neurons, background spike sources or are injected from off-chip and routed in-between on-chip and off-chip entities.
\Cref{fig:methods:chipfoto} shows an annotated image of a \gls{bss2} chip.

In addition, two embedded \gls{simd} microprocessors have access to the chip configuration and to observables of the neural network core allowing for plasticity implementations including the computation of decisions and modifications of the chip configuration during experiment runtime.
They feature \SI{16}{\kibi\byte} on-chip \gls{sram} memory and implement a subset of the Power~\gls{isa} 2.06~\cite{powerisa_206} with a custom vector extension including \SI{8}{\bit} and \SI{16}{\bit} unsigned integer and signed saturating fractional arithmetics on \SI{128}{\byte} vector data~\cite{friedmann2016hybridlearning}.
The vector units of the processors can directly access synapse and neuron observables (weights, spike routing configuration and \gls{adc} measurements of synapse-local causal and anticausal accumulated correlation data, as well as membrane, adaptation and synaptic inputs potentials of the neurons) in parallel.
Other configuration and observables are accessible via the memory-mapped \SI{32}{\bit} Omnibus~\cite{githubomnibus} using the scalar unit of the processors.
This allows for arbitrary changes to the neural network's topology and parameterization and notably includes access to per-neuron spike event counter values.
However, direct access to spike event timing data during the experiment is currently not possible.\footnote{In principle, recorded event data could be made available via off-chip memory accesses in the future.}

The experiment is controlled in real time by a sequential state machine on a dedicated \gls{fpga}, which is programmed by a host computer connected via \gls{gbe}.
The \gls{fpga} is connected via eight \SI{1}{\giga\bit\per\second} links to the chip, transporting events and memory transfers.
It exposes \gls{bram} (currently \SI{128}{\kibi\byte}) and attached \gls{dram} (\SI{256}{\mebi\byte}) to both embedded processors via the \gls{fpga}-chip links.

We build upon the software stack introduced in~\cite{mueller2022scalable}.
Our user front end experiment notation including plasticity targets PyNN~\cite{davison2009pynn}.
It allows abstract network descriptions without requiring placement information for the hardware back end by the user.
Additionally, its target audience are computational neuroscience experiments, which are currently the primary experiments implementing online plasticity.
Network topology is described by constructing populations of neurons and projections of synapses in-between.
Plastic network components for other PyNN back ends like Brian~2~\cite{stimberg2019brian} are implemented as synapse types, which are homogeneous across a projection, e.g.\ for \gls{stdp}.

The PyNN front end implementation for \gls{bss2} makes use of a signal-flow graph-based back end\footnote{\url{https://github.com/electronicvisions/grenade}}, which first represents the unplaced network topology, then performs the mapping to hardware components (synapse and neuron circuits as well as event routing entity configuration) and constructs a transformed graph including used hardware components and their signal flow.
This mapped experiment description is then \gls{jit} compiled and executed on the hardware.
The gray part of \cref{fig:results:execution_model} shows this execution flow.

Programs for the embedded processors are cross-compiled using the toolchain introduced in~\cite{mueller2020bss2ll} written in \texttt{C++} with limited standard library support, a cross-compiled hardware abstraction layer for accessing the neural network core~\cite{mueller2022scalable} as well as a support library for processor-specific functionality\footnote{\url{https://github.com/electronicvisions/libnux}} like accessing the synapse array with the \gls{simd} unit.
Since there's currently no general-purpose operating system available on the embedded processors, all programs are monolithic and freestanding.

\section{Results}\label{sec:results}
We will now look at the execution model for experiments on \gls{bss2}, the data flow of the experiments, the associated user interface and the generation of code for the programmable plasticity.
Finally, we evaluate the performance of our implementation.
\subsection{Execution model}
We implement plasticity rules as parts of the program to run on the embedded processors.
The complete program is \gls{jit} compiled from the provided user input prior to execution of the experiment.
The embedded processor program, the experiment on the neural network core, as well as the \gls{fpga} run concurrently.
Therefore, synchronization is required.
For this purpose, the embedded processor program waits for a start command from the \gls{fpga}, which synchronizes its program execution with the \emph{experiment time}.

\begin{figure}[htbp]
\includegraphics[width=0.48\textwidth]{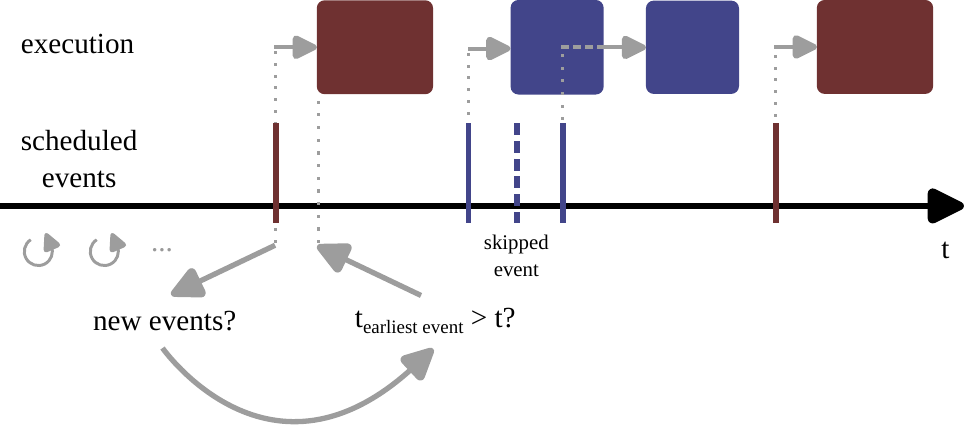}
\caption{\label{fig:results:scheduling_loop}
Scheduling of plasticity rules on the embedded processors.
Different colors represent different plasticity rules.
The program loops and fetches new events marking plasticity rules to be executed.
It then checks whether the earliest event deadline has passed and issues the execution of the corresponding plasticity rule.
The arrows represent the latency incurred between the event's deadline and the beginning of the execution of the associated rule.
Events, for which the deadline passes during another execution are skipped (dashed event), except for the latest one, for which execution is additionally delayed by the amount of time passed in the execution of the current rule.
}
\end{figure}

In order to support multiple plasticity rules during an experiment, a scheduler running on the embedded processors is used for their execution.
Since low computational overhead is required, we considered a pre-computed event list which is only iterated or an earliest-deadline-first scheduler.
While the pre-computed event list is expected to show the smallest runtime overhead, it is limited in terms of the number of events by memory constraints and doesn't support dynamic creation, deletion, or modification of events (to be generated) during experiment runtime.
Therefore, we choose an earliest-deadline-first scheduler as our default.
\Cref{fig:results:scheduling_loop} depicts the scheduling flow of plasticity rules---or other tasks---on the embedded processors.

We expect each plasticity rule to adhere to a defined callable interface, see~\cref{lst:results:kernel_interface}, which is invoked by the scheduler.
The points in time, during the experiment runtime, at which the plasticity rule shall be executed are fed into the scheduler during the experiment.

Both embedded processors share the same program and are made aware of their location on the chip, allowing them to take different code paths without incurring the potentially unnecessary execution of plasticity rules.
This allows more memory to be available for instructions, since the shared memory accessible via the \gls{fpga} does not need to be statically divided and inter-processor communication is possible via shared memory.

\begin{figure}[htbp]
\includegraphics[width=0.48\textwidth]{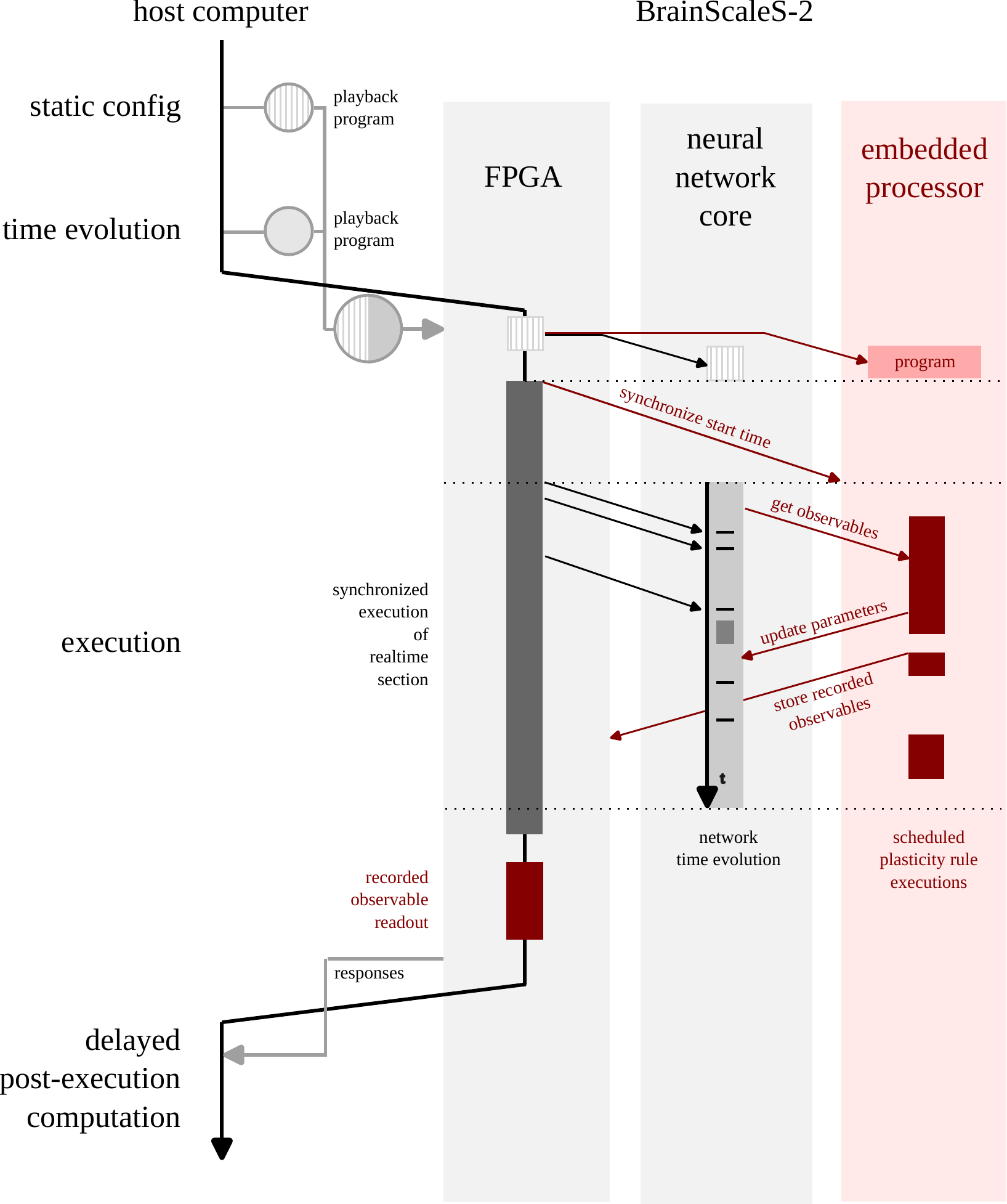}
\caption{\label{fig:results:execution_model}
Execution model for experiments on \gls{bss2}.
The developed support for plasticity on the embedded processors is depicted in red, the previously implemented execution model including the host computer, \gls{fpga}, and neural network core on \gls{bss2} is depicted in gray.
Playback programs for the \gls{fpga} for static configuration of the system and subsequent neural network time evolution are generated on the host computer.
Execution on the \gls{fpga} is synchronized with the neural network core time evolution, e.g., by injecting external spike events.
After execution, the responses from the system are read out to the host computer, post-processed, and provided to the user in PyNN.
For plasticity, the embedded processors are provided with their program during the static configuration phase of the experiment.
Concurrent with the time-continuous evolution of the analog neural network core, it accesses observables, updates network parameters and stores to-be-recorded observables.
The scheduling of the plasticity rules on the embedded processors is time-synchronized at the beginning of the real-time section of the experiment by the \gls{fpga}.
After completion of the concurrent real-time experiment execution, the recorded observable data and other responses are read out and made available to the user after post-processing.
}
\end{figure}

\Cref{fig:results:execution_model} visualizes the implemented execution model including plasticity: this work implemented the red components.

\subsection{Data flow}

Experiments on neuromorphic hardware often require access to observables during or after execution.
In addition to spikes, neuronal or synaptic dynamics, in experiments with plasticity the evolution of the configuration of altered network components is of interest.
During experiment design, plasticity-rule-internal variables might also be useful.
Therefore, to-be-recorded observables need to be user-specifiable for a maximal flexibility in plasticity design.
Since there's no dedicated hardware mechanism for recording such observables, we store them to memory, which is directly reachable by the embedded processors during experiment runtime.
After the completion of the experiment, this memory is read out from the host computer for further processing.
In order to support long-running experiments with many recorded observables, we choose to use the \gls{dram} attached to the FPGA as intermediate observable storage.

Since the architecture of the host computer (typically \SI{64}{\bit} x86) and the embedded processors (\SI{32}{\bit} Power) differ, platform-agnostic serialization of the recorded data is required.
We optimize for low required compute power and memory footprint on the serializing side, i.e.\ the embedded processors, and accept imperfect deserializing performance on the host computer, since the embedded processor serialization is real-time-critical during experiment runtime.
Therefore, we directly store the native data structures to memory (by placing global entities of their type in the \gls{dram} memory, which allows for all the typical \texttt{C++} language features for handling the data without manual casting or copying) and deserialize their memory layout manually on the host computer.
In the future, low-footprint cross-platform serialization libraries like \texttt{bitsery}~\cite{vinkelis2020bitsery} would resolve this need for hand-crafted deserialization, which relies on the stable memory layout of a fixed set of data types.

\subsection{User interface}

Users formulate experiments on the neuromorphic hardware by specifying the network topology and experiment dynamics in a high-level language.
For experiments involving plasticity, we use PyNN~\cite{davison2009pynn}.
To support multi-factor rules, multiple network entities can be subject to a plasticity rule.
Therefore, in contrast to PyNN's standard synapse types including fixed plasticity rules (e.g., the \gls{stdp} mechanism), we separate the plasticity rule definition from the other network entities and assign network entities to be accessed by the plasticity rule separately.
A base class is provided for plasticity rules, which users can derive from and implement missing functionality.
After creation of a plasticity rule instance, it can be assigned to be the plasticity rule of a neuron-cell type of a population, or a synapse type of a projection.
This in turn gives it access to the configuration and observables of that network entity.
\Cref{lst:results:pynn_interface} shows this interface.
\begin{listing}[htbp]
\begin{minted}{python}
import pynn_brainscales.brainscales2 as pynn

class MyPlasticity(pynn.PlasticityRule):
  ...

my_plasticity = MyPlasticity(...)

pop = pynn.Population(
  n,
  pynn.standardmodels
  .cells.CalibratedCubaNeuron(
    plasticity_rule=my_plasticity))

proj = pynn.Projection(
  pop,
  pop,
  pynn.AllToAllConnector(),
  pynn.standardmodels
  .synapses.PlasticSynapse(
    plasticity_rule=my_plasticity))
\end{minted}
\caption{\label{lst:results:pynn_interface}
	Example of plasticity rule creation and assignment of accessible network components in PyNN.
	After the creation of the plasticity rule, by assigning it as the plasticity rule of the population's cell type and the projection's synapse type, they are made accessible to the plasticity rule kernel.
}
\end{listing}

The derived plasticity rule includes execution timing information as well as observable recording information.
Users can specify the times during the experiment runtime when the plasticity rule shall be executed.
Currently, only periodic timing is supported: it includes one-shot execution and is formulated as a generator, which yields constant memory requirements independent of the experiment runtime.
Due to the variety of observables of interest in plasticity rules, users can specify an arbitrary number of named observables to record for each execution of the plasticity rule.
Currently, signed and unsigned \SI{8}{\bit} and \SI{16}{\bit} integer values per synapse in a projection or per neuron in a population are supported.
To allow users to optimize for space or execution runtime, the choice of a packed (smaller) or unpacked row-wise (faster) recording storage is possible, since the latter can directly be written by the vector unit of the embedded processors, while the former requires looping the entries via the scalar unit.
Depending on the number of used neuron circuits in a population or synapse columns in a projection, this choice incurs no memory overhead (if all columns are used) or up to \num{256} times the actual memory used (if only one entry is of interest).
\Cref{lst:results:observable_recording} shows this choice and the resulting recording storage structure.
\begin{listing}[htbp]
\begin{minted}{python}
observables = {
  "my_obsv": PlasticityRule.ObservablePerSynapse(
    uint8, LayoutPerRow.packed),
  ...
}
\end{minted}
\begin{minted}{cpp}
struct Recording {
  tuple<
    array<
      array<uint8_t, num_columns>,
      num_rows>,
    ...> my_obsv;

  ...
};
\end{minted}
\caption{\label{lst:results:observable_recording}
	Example of user-specified observables for the plasticity rule (top) and the corresponding generated recording data structure available in the user-provided plasticity function (bottom).
	The chosen name of the observable is directly translated to a member of the recording data structure, which in the case of one observable per synapse is available for each projection (entry in \mintinline{cpp}|tuple|) in row-major array order.
	Here, packed storage is used, where for each synapse one data entry is available.
	In unpacked storage, the inner \mintinline{cpp}|array<uint8_t, num_columns>| is replaced by a \gls{simd} vector containing entries for all (even unused) columns of a synapse row on the chip.
}
\end{listing}

The main user input to the plasticity rule is its ``kernel'' code.
Here, the user directly provides the \texttt{C++} code of the plasticity rule to be compiled for the embedded processors.
For each plasticity rule, we require a functional interface, which is parameterized by handles to the accessible network entities, synapses or neurons, as well as a reference to the observable recording memory block per execution of the rule.
In order to enable the implementation of plasticity rules with state beyond their accessible network entity configuration such as weights, global or rule-instance-local variables can be employed.
\Cref{lst:results:kernel_interface} shows the kernel function interface.
\begin{listing}[htbp]
\begin{minted}{cpp}

T global_state_variable;
static T local_state_variable;

void PLASTICITY_RULE_KERNEL(
  array<SynapseArrayViewHandle, N> const& synapses,
  array<NeuronViewHandle, M> const& neurons,
  Recording& recording)
{
  ... global_state_variable ...
  ... local_state_variable ...
}
\end{minted}
\caption{\label{lst:results:kernel_interface}
	Example of plasticity rule kernel code supplied to the PyNN plasticity rule.
	The kernel function receives handles to synapses of projections and neurons of populations accessible to the plasticity rule, as well as a reference to a memory block for the recording of user-defined observables.
	Global or plasticity-rule-local state is achieved via global or static variables used in the kernel function.
}
\end{listing}

To minimize the need for detailed chip knowledge, no hardware substrate-specific information, such as the placement of neurons on the chip, needs to be specified.
However, during the experiment runtime, the plasticity rule requires this placement information to access the correct hardware entities associated with the network elements in the user-defined topology.
We supply this information in the synapse and neuron handles passed as input parameters to the kernel function during runtime.
The user therefore can assume when writing the plasticity code, that this information will be provided at the invocation of the rule function.
For this, code generation is used to embed the placement information at compile time into the program for the embedded processors.

\subsection{Code generation}

We employ automated code generation for program components for the embedded processors which are known prior to experiment execution.
This enables composability of multiple rules to be part of one program and the injection of automatically generated information.
We embed all static information known prior to experiment execution at compile time of the experiment to optimize runtime performance.
After the complete experiment description (including plasticity) is provided by the user, the scheduling requirements for the plasticity kernels, the expected recorded observables, and the network components to which the plasticity rules shall be applied can be computed.
From this, we generate code for user-specified observable recording data storage structures (cf.\ \cref{lst:results:observable_recording}), injection of supplied plasticity rules into the execution scheduling and provisioning of post-hardware-mapping location information of accessible synapses and neurons to the plasticity functions.
We use \texttt{inja}\footnote{\url{https://github.com/pantor/inja}}, a template engine for \texttt{C++}, for generation of the code for the embedded processor.

\subsection{Evaluation}\label{sec:results:evaluation}

\begin{figure*}[htbp]
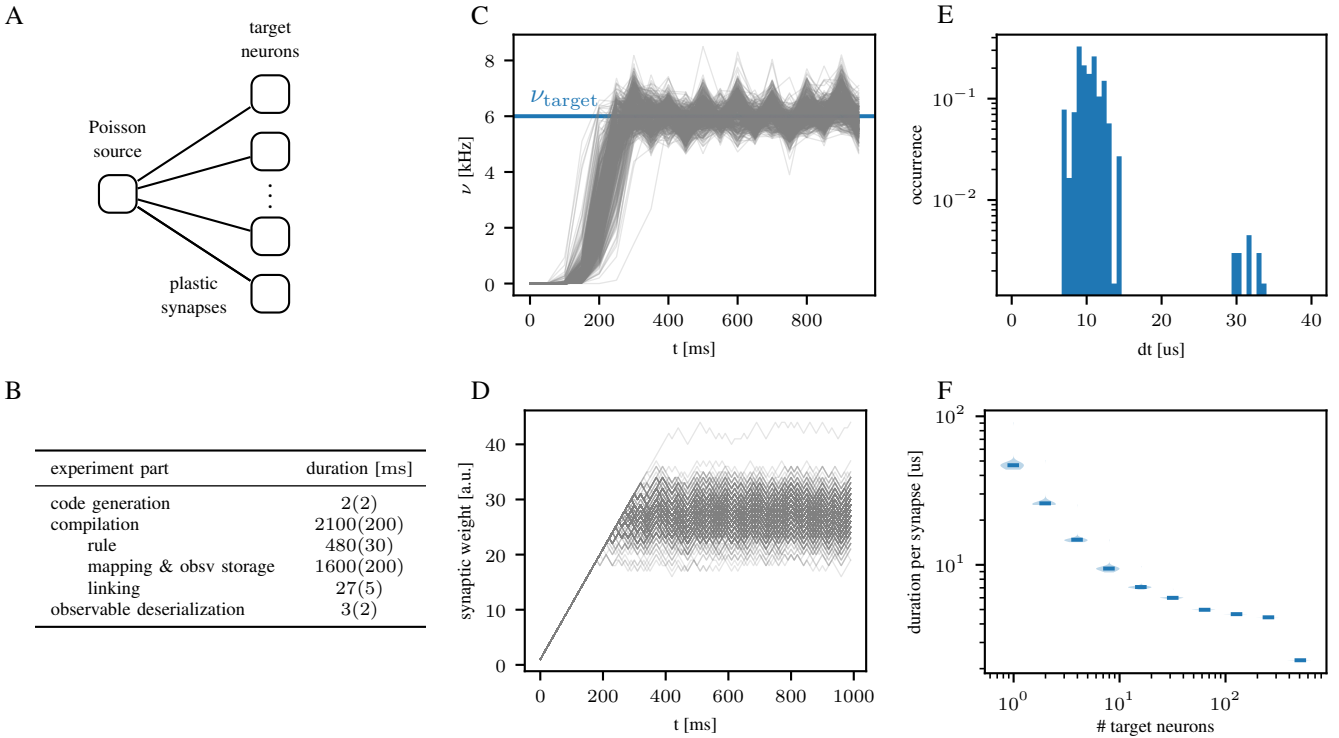


    \tikzset{
             panel/.style={
                     inner sep=0pt, outer sep=0, execute at begin node={\tikzset{anchor=center, inner sep=.33333em}}
             },
                label/.style={anchor=north west, inner sep=0, outer sep=0}
        }
\begin{tikzpicture}
	\node[panel, anchor=north west] (a) at (0, 0) {
	};
	\node[thick, rounded corners, draw, rectangle, anchor=center, minimum size=0.5cm] (cs) at (1.5, -2.5) {};
	\node[anchor=center, above=0.1 of cs, align=center] (cst) {\scriptsize{Poisson}\\[-0.3em]\scriptsize{source}};
	\node[thick, rounded corners, minimum size=0.5cm, right=1.5 of cs, rotate=90, anchor=north] (cd) {\dots};
	\node[thick, rounded corners, minimum size=0.5cm, draw, rectangle, anchor=center, above=0.3 of cd.center] (cnt1) {};
	\node[thick, rounded corners, minimum size=0.5cm, draw, rectangle, anchor=center, above=0.5 of cnt1.center] (cnt2) {};
	\node[anchor=center, above=0.1 of cnt2, align=center] (cnt) {\scriptsize{target}\\[-0.3em]\scriptsize{neurons}};
	\node[thick, rounded corners, minimum size=0.5cm, draw, rectangle, anchor=center, below=0.3 of cd.center] (cnb1) {};
	\node[thick, rounded corners, minimum size=0.5cm, draw, rectangle, anchor=center, below=0.5 of cnb1.center] (cnb2) {};
	\draw[thick] (cs) -- (cnt1) {};
	\draw[thick] (cs) -- (cnt2) {};
	\draw[thick] (cs) -- (cnb1) {};
	\draw[thick] (cs) -- (cnb2) {};
	\draw[thick] (cs) to [edge node={node[below=0.3, align=center] {\scriptsize plastic\\[-0.3em]\scriptsize synapses}}] (cnb2) {};
	\node[label] at (a.north west) {A};

	\node[panel, anchor=north west] (b) at (0.4, -5.9) {
		\scriptsize
		\begin{tabular}{lc}
			\toprule
			experiment part & \textrm{duration} [\textrm{\si{\ms}}]\\
			\midrule
			code generation & \num{2\pm2}\\
			compilation & \num{2100\pm200}\\
			\qquad rule & \num{480\pm30} \\
			\qquad mapping \& obsv storage & \num{1600\pm200} \\
			\qquad linking & \num{27\pm5}\\
			observable deserialization & \num{3\pm2}\\
			\bottomrule
		\end{tabular}
	};
	\node[label] at (0, -5) {B};

	\node[panel, anchor=north west] (c) at (5.6, 0) {
		\input{plot_plasticity_rule_homeostasis_512.pgf}
	};
	\node[label] at (6.2, 0) {C};

	\node[panel, anchor=north west] (d) at (5.6, -5) {
		\input{plot_plasticity_rule_homeostasis_weights_512.pgf}
	};
	\node[label] at (6.2, -5) {D};

	\node[panel, anchor=north west] (e) at (11.6, 0) {
		\input{plot_plasticity_rule_homeostasis_time_jitter.pgf}
	};
	\node[label] at (12.4, 0) {E};

	\node[panel, anchor=north west] (f) at (11.6, -5) {
		\input{plot_plasticity_rule_homeostasis_time_duration_rel.pgf}
	};
	\node[label] at (12.4, -5) {F};
\end{tikzpicture}
\caption{\label{fig:results:evaluation_homeostasis}
	Evaluation of an exemplary homeostatic plasticity rule.
	A Poisson source of \SI{120}{\kilo\Hz} (hardware time domain) projects via plastic synapses onto a variable number of target neurons, see (A).
	(B) shows the duration of compilation and code generation for the programs executed on the embedded processors.
	(C) shows the firing rate of the neurons (for 512 target neurons).
	The homeostatic plasticity is successful in converging the neurons' firing rate to the target firing rate of \SI{6}{\kilo\Hz} (depicted in blue).
	Firing rates are extracted from the recorded spike trains by counting in bins of \SI{50}{\ms}.
	Due to the source being Poisson, no perfect convergence to the target firing rate is expected.
	(D) shows the evolution of the synaptic weights.
	(E) shows the difference of the actual rule execution times to the expected/scheduled execution times.
	Due to using the earliest-deadline-first scheduler for rule execution, a constant positive shift of the actual to the expected execution time is visible.
	The occurrence of the higher difference is due to the instruction caches of the embedded processors being cold at the first rule execution.
	(F) shows the rule execution duration per plastic synapse in dependence of the number of target neurons.
	Here, the usage of the embedded processors \gls{simd} unit shows runtime performance advantages for a higher number of target neurons.
	Between \num{512} and \num{256} target neurons a factor of two in runtime per plastic synapse is achieved due to both embedded processors executing the rule for their hemisphere in parallel for \num{512} target neurons, while for \num{256} only one processor is active.
}
\end{figure*}
\begin{table}
\caption{\label{tab:results:evaluation_env}
The toolchain and compute environment used for the evaluation experiments described in~\cref{sec:results:evaluation}.
}
\begin{minipage}{\textwidth}
\begin{tabular}{ll}
\toprule
Component & Description \\
\midrule
Host computer &\\
\qquad \Gls{cpu} & two-socket AMD Epyc 7543\\
\qquad Main memory & \SI{1}{\tebi\byte}\\
Toolchain for embedded processors & \\
\qquad Compiler & \texttt{gcc}\footnote{\url{https://github.com/electronicvisions/gcc}} 8.1.0 (ext'd for \acrshort{bss2}) \\
\qquad Linker & \texttt{bfd} from \texttt{binutils}\footnote{\url{https://github.com/electronicvisions/binutils-gdb}} 2.38\\
\qquad \texttt{libc} & \texttt{newlib}\footnote{\url{https://github.com/electronicvisions/newlib}}@9c84bfd4\\
\bottomrule
\end{tabular}
\end{minipage}
\end{table}

Here, we evaluate a local homeostatic plasticity rule utilizing access to neuron firing rates as an exemplary use case to evaluate the real-world applicability of our approach.
The experiment is also available for execution via the EBRAINS research infrastructure.\footnote{\url{https://github.com/electronicvisions/brainscales2-demos/blob/master/ts_11-plasticity_homeostasis.rst}}
The learning rule weight updates are calculated according to:
\begin{equation}
	\Delta w = \text{sign}\left(\nu_\text{target} - \nu \right),
\end{equation}
where we choose the $\text{sign}$ function for constant weight updates, which are not subject to resolution limits when approaching the target firing rate while exhibiting potentially unstable behavior after convergence.
\Cref{fig:results:evaluation_homeostasis} shows the experiment and its results.
We evaluate the correctness of the solution and runtime as well as its compile-time performance metrics.
\Cref{tab:results:evaluation_env} lists information about the toolchain and compute environment used.
The plasticity rule is correctly applied such that the neurons fire according to the expected mean target firing rate.
We achieve an execution timing accuracy of \SI{10\pm3}{\us}, cf.\ \cref{fig:results:evaluation_homeostasis}{E}.
The rule execution duration improves for more target neurons, as the \gls{simd} unit used for calculating the weight updates has a higher utilization.
At best, \SI{2.28\pm0.06}{\us} per plastic synapse is reached.
The compilation time is measured to be \SI{2.1\pm0.1}{\s}, while the code generation time of \SI{2\pm2}{\ms} and observable deserialization time of \SI{3\pm2}{\ms} are insignificant in comparison.
All three increase with a larger amount of target neurons, since providing location information for more neurons requires more generated and compiled code, while also requiring a larger amount of data to be deserialized for the recorded observables.

For an increased number of plasticity rules to be executed, we expect the compilation and code generation duration as well as the execution time offset to scale linearly.
The rule execution durations are expected to remain unaffected except for more potential instruction cache misses and the necessity to fetch more instructions from memory instead of using the cache due to more different code sections being executed.

Additionally, runtime memory resources are limited for both data and instructions and therefore considered.
To limit the required memory for storing projection location information on the embedded processors, only (possibly strided) rectangle-shaped projections are supported, since for them, only the rows and columns are required to be provided to the program during runtime.
The memory footprint therefore scales with
\begin{equation}
M \propto \sum_{j}^\text{projections} \left( \text{\#columns}_j + \text{\#rows}_j \right) + \sum_{j}^\text{populations} \text{\#neurons}_j,
\end{equation}
which favors---from a single-chip point of view---coarse-granular network topologies.
Observable storage scales depending on whether the information is stored as packed array (slow access via scalar unit, memory-efficient) or in full rows (fast access via \gls{simd} unit, potentially memory-inefficient).
In the latter case the memory usage for each row is rounded up to \num{256} entries, where in the former case the number of entries per row is exactly equal to the used columns for projections or neurons for populations.
Memory usage per rule invocation then scales with:
\begin{align}
M \propto & \sum_{o}^\text{syn obsv} \sum_{j}^\text{projections} \text{\#columns}_j \cdot \text{\#rows}_j \nonumber \\
+& \sum_{o}^\text{nrn obsv} \sum_{j}^\text{populations} \text{\#neurons}_j,
\end{align}
where $\text{\#columns}$ and $\text{\#neurons}$ might be rounded up to \num{256} if they are stored unpacked for direct \gls{simd} unit access.

\section{Discussion}\label{sec:discussion}

In this work, we facilitate the use of the \gls{bss2} system as a time-continuous spiking neural network accelerator with programmable plasticity.
While the analog neural network core provides a substrate for fast emulation of complex neuronal morphologies, the processors are programmable and support various forms of plasticity~\cite{friedmann2016hybridlearning}.
Based on the backend-agnostic \gls{snn} description language PyNN~\cite{davison2009pynn}, our software supports the unified definition of \gls{snn} topology and experiment protocols, as well as plasticity rules, thereby simplifying the application of plasticity in the \gls{bss2} neuromorphic systems.

We provide an interface for the description of local and non-local plasticity rules and provide access to customizable observable recording.
Users formulate plasticity rules by supplying the code for the embedded processors, their execution timing, and the observables to be captured, and describe the application of the rule to network entities in PyNN.
Prior to execution on the hardware, the plasticity rule code for the embedded processors is \gls{jit}-compiled.
This allows for automated placement of network entities to hardware components and annotation of this information to the compiled programs without user interaction or knowledge of the placement result.
During the experiment, the embedded processor program is executed synchronously with the analog network emulation.
Recorded observables are provided in PyNN data structures after experiment execution.
The presented plasticity integration and abstraction is already used in~\cite{dietrich2023multi}, where they use non-local \gls{stdp}-based plasticity for sequence learning, and in~\cite{atoui2024multi} where a complex calcium-based plasticity rule is implemented on \gls{bss2}.
Here, we evaluate the successful application of a homeostatic rule onto target neurons as an artificial use case.
The analyzed compile-time and runtime performance metrics verify the expected scaling behavior.

However, several limitations remain for this ``hybrid'' approach to plasticity:
First, the continuous-time dynamics of the analog core and the finite processor speed limit the complexity of plasticity rules with strict temporal requirements.
Second, efficiency gains by the processor's vector unit are only achieved in cases when many synapses with the same plasticity rules are processed simultaneously;
if many different rules are used, or if the placement of plastic synapses is heterogeneous, the vector unit cannot be used efficiently, which could be mitigated by future improvements to the \gls{simd} unit, such as vector permutation and packing instructions, and support for reduced-width vector registers.
Third, in the existing implementation, modelers are relieved of the manual ``placement'' of plasticity rules and of the peculiarities of data exchange, but there is still a need to write relatively low-level code:
plasticity kernels require familiarity with specific functions for observing and modifying parameters, as we have yet to develop a domain-specific language.\footnote{For numerical simulation, there are languages such as LEMS~\cite{cannon2014lems} or integrated support as in Brian~2~\cite{stimberg2019brian}.}

While the presented plasticity integration targets PyNN, it can easily be adapted for other front ends, as the middleware implementation in the graph-based experiment notation can be shared between front ends.
Integration with an event-based machine training framework focusing on \glspl{snn}, cf.\ \texttt{jaxsnn}~\cite{mueller2024jaxsnn}, would provide a natural way to combine gradient-based training and online plasticity, or meta learning, when gradient-based training is applied to hyperparameters of the plasticity rules.

\section*{Acknowledgements}
\addcontentsline{toc}{section}{Acknowledgment}

This work has received funding from
the EC Horizon 2020 Framework Programme
under grant agreements
945539 (HBP SGA3), 
the EC Horizon Europe Framework Programme
under grant agreement
101147319 (EBRAINS 2.0),
the \foreignlanguage{ngerman}{Deutsche Forschungsgemeinschaft} (DFG, German Research Foundation) under Germany's Excellence Strategy EX 2181/1-390900948 (the Heidelberg STRUCTURES Excellence Cluster).

\section*{Author Contributions}\label{sec:author_contributions}

We give contributions in the \textit{CRediT} (Contributor Roles Taxonomy) format:
\textbf{PS}: Conceptualization, visualization, methodology, software, resources, writing — original draft, writing — reviewing \& editing;
\textbf{EM}: Conceptualization, methodology, software, resources, writing — original draft, writing — reviewing \& editing, supervision, funding acquisition;
\textbf{JS}: Supervision, funding acquisition, writing — reviewing \& editing.

\printbibliography

\end{document}